\documentclass[10pt,twocolumn,letterpaper]{article}

\usepackage[pagenumbers]{cvpr}
\usepackage{graphicx}
\usepackage{pifont} 
\usepackage{multirow}
\usepackage{xspace}
\usepackage{booktabs}

\newcommand{\ours}{\textsc{Aurora}\xspace}

\makeatletter
\renewcommand{\paragraph}{%
  \@startsection{paragraph}{4}{\z@}%
  {1.25ex \@plus 1ex \@minus .2ex}
  {-1em}
  {\normalfont\normalsize\bfseries}%
}
\makeatother

\usepackage{tabu}
\usepackage{xcolor}

\definecolor{cvprblue}{rgb}{0.21,0.49,0.74}
\usepackage[pagebackref,breaklinks,colorlinks,allcolors=cvprblue]
{hyperref}

\newcommand{\model}[0]{\textsc{Aurora}\xspace}
\newcommand{\crossmark}{\ding{55}} 
\newcommand{\ourcheckmark}{\ding{51}} 
\definecolor{dartmouthgreen}{rgb}{0.05, 0.5, 0.06}
\definecolor{deepcarmine}{rgb}{0.66, 0.13, 0.24}
\def\paperID{12371} 
\def\confName{CVPR}
\def\confYear{2025}

\title{Perception Tokens Enhance Visual Reasoning in Multimodal Language Models}

\author{%
  \bf Mahtab Bigverdi$^{1}$  \: \:
  \bf Zelun Luo$^{2}$  \: \:\bf
  Cheng-Yu Hsieh$^{1}$  \: \: \\\bf
  Ethan Shen $^{1}$ \: \:
  Dongping Chen $^{1}$ \: \:
  Linda G. Shapiro$^{1}$ \: \: Ranjay Krishna$^{1}$\\
  $^1$University of Washington, 
  $^2$Google Research\\
  \texttt{\{mahtab,cydhsieh,ethans03,shapiro,ranjay\}@cs.washington.edu,}\\  \texttt{alanzluo@gmail.com}, \texttt{cdp0612@uw.edu}
}

\begin{document}
\maketitle

\begin{abstract}

Multimodal language models (MLMs) still face challenges in fundamental visual perception tasks where specialized models excel. Tasks requiring reasoning about 3D structures benefit from depth estimation, and reasoning about 2D object instances benefits from object detection. Yet, MLMs can not produce intermediate depth or boxes to reason over.
Finetuning MLMs on relevant data doesn't generalize well and outsourcing computation to specialized vision tools is too compute-intensive and memory-inefficient.
To address this, we introduce \textbf{Perception Tokens}, intrinsic image representations designed to assist reasoning tasks where language is insufficient. Perception tokens act as auxiliary reasoning tokens, akin to chain-of-thought prompts in language models. For example, in a depth-related task, an MLM augmented with perception tokens can reason by generating a depth map as tokens, enabling it to solve the problem effectively.
We propose \ours, a training method that augments MLMs with perception tokens for improved reasoning over visual inputs. \ours leverages a VQVAE to transform intermediate image representations, such as depth maps into a tokenized format and bounding box tokens, which is then used in a multi-task training framework.
\ours achieves notable improvements across counting benchmarks: $+10.8\%$ on BLINK, $+11.3\%$ on CVBench, and $+8.3\%$ on SEED-Bench, outperforming finetuning approaches in generalization across datasets. It also improves on relative depth: over $+6\%$ on BLINK.
With perception tokens, \ours expands the scope of MLMs beyond language-based reasoning, paving the way for more effective visual reasoning capabilities. Code will be released at the \href{https://aurora-perception.github.io/}{project page}.

\end{abstract}    

\begin{figure}[t]
    \centering
    \includegraphics[width=\linewidth]{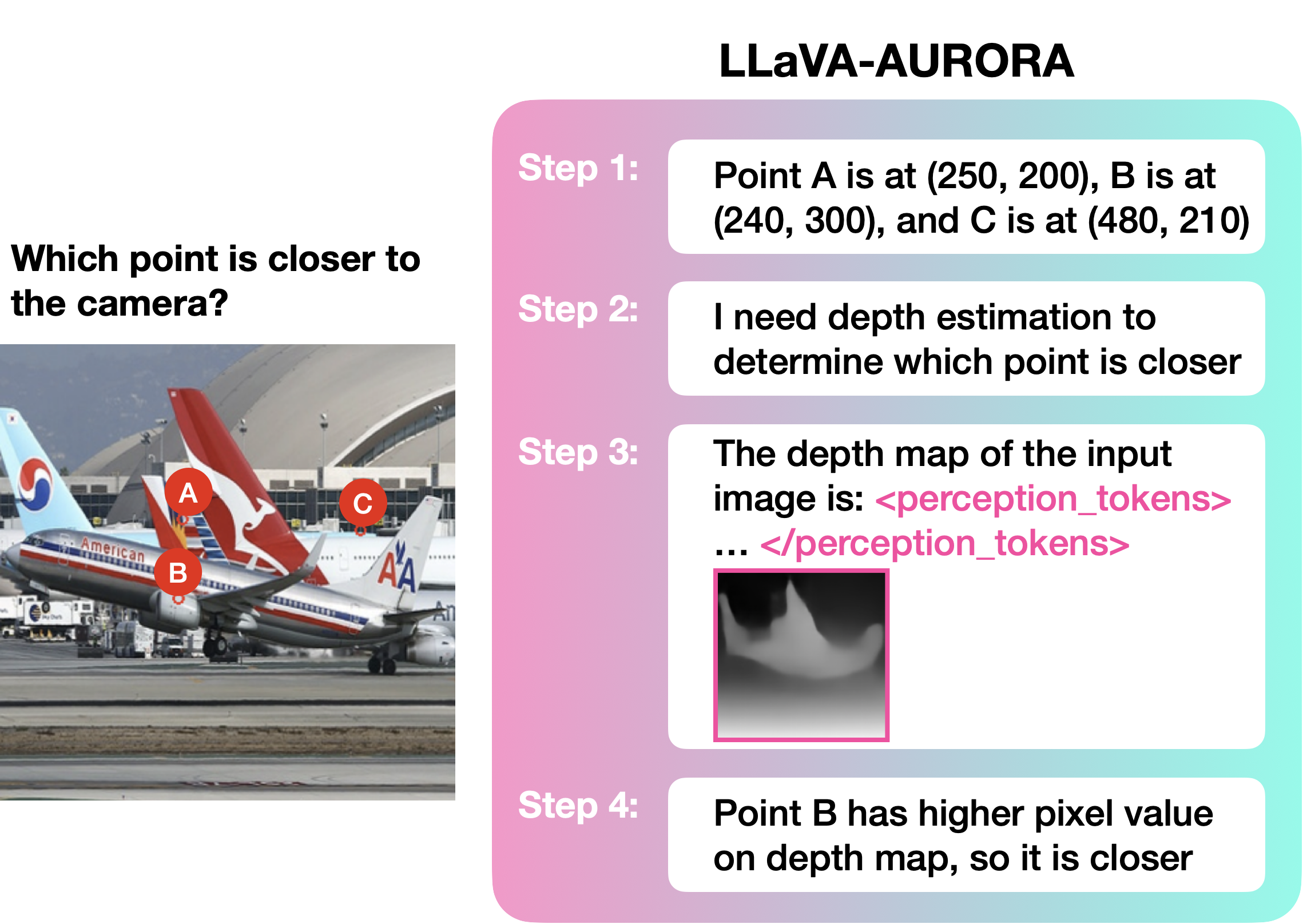}
    \vspace{-6mm}
    \caption{We introduce Perception Tokens, intermediate reasoning tokens that allow MLMs to go beyond using language in reasoning. With it, we develop \ours, a framework that trains multimodal language models to leverage visual perception tokens, allowing them to use depth estimation and bounding box predictions while reasoning.}
    \label{fig:teaser}
    \vspace{-4mm}
\end{figure}

\section{Introduction}
\label{sec:intro}

\begin{figure*}[t]
    \centering
    \includegraphics[width=\textwidth]{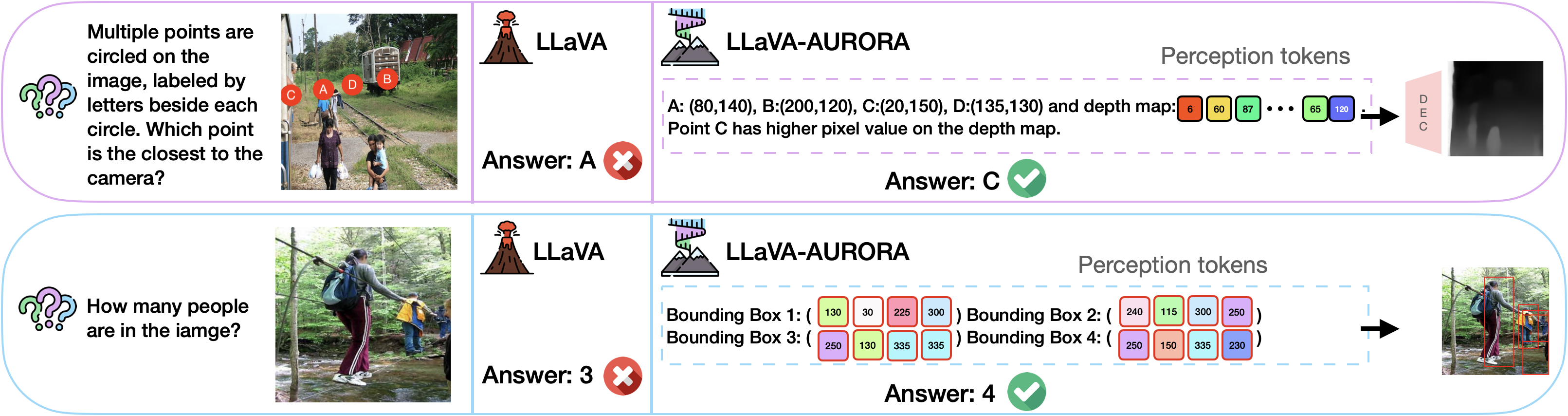}
    \vspace{-8mm}
    \caption{We demonstrate relative depth estimation and counting questions where LLaVA fails. In contrast, by learning to utilize visual perception tokens as intermediate reasoning steps, LLaVA-\ours successfully complete these tasks requiring perceptual understanding.
    }
    \vspace{-3mm}
    \label{fig:image_figure}
\end{figure*}

In contrast to the growing emphasis on building multimodal language models (MLMs), computer vision was originally attempting to interpret images as projections of indescribable 3D intrinsics, not just processing 2D arrays of language patterns~\cite{do1961machine,minsky1969introduction,marr2010vision}.
Towards this endeavor, early vision research developed a series of intermediate image representations—enabling geometric reasoning through depth estimation~\cite{torralba2002depth} and instance reasoning through bounding box grounding~\cite{harris1988combined}.
As pointed out by recent work, we have focused less on such perceptual representations and instead tackled reasoning problems that require limited visual involvement~\cite{fu2024blink,tong2024cambrian1,tong2024eyes}.
This is likely because many traditional vision tasks remain ambiguous through natural language. Consider the task of identifying which of a set of N points is furthest away from the camera. While language doesn't lend itself to reason over this problem, a depth estimation would provide the appropriate abstraction to reason over.

Numerous attempts have been made to enable MLMs to reason over intrinsic image representations.
The default approach is to finetune MLMs on data tailored to the specific perception task of interest, hoping that the model implicitly learns the required intrinsic representations~\cite{chen2024spatialvlm}. 
Another option is to outsource the computation to external tools: the MLM can invoke a depth estimator or object detector to produce the appropriate intrinsic~\cite{hu2024sketch}.
Unfortunately, relying on external models makes the task more computationally expensive and requires loading additional models with more memory. Similarly, vanilla fine-tuning (even with advancements like LoRA~\cite{hu2021lora}) has shown marginal improvements.

Conceptually, we introduce \textbf{Perception Tokens}, intrinsic image representations that aid in reasoning where language is insufficient.
To solve the aforementioned task, an MLM augmented with perception tokens can solve the task similar to how language models use chain-of-thought. 
They will produce a response like the following: ``\textit{The depth map is $<$perception tokens$>$. Therefore, point D is closest to the camera.}'' 
Here, \textit{$<$perception tokens$>$} is a set of tokens that implicitly estimate the depth of the image.
Similarly, for a counting task, the model can first generate perception tokens that represents the location of the relevant bounding boxes of the desired object, and count the number of boxes to support its final answer.

To demonstrate the utility of perception tokens, we introduce \ours, a training algorithm to augment MLMs with the ability to use perception tokens as intermediate reasoning steps. 
For certain intermediate representations (\eg depth maps), we train a VQVAE to transform them into a set of tokens, treating the learned VQVAE codebook indices as a collection of perception tokens. while for others, such as bounding boxes, we use directly encoded structured tokens.
Next, we follow a multi-task training approach~\cite{hu2024visual,hsieh2023distilling} to train MLMs to use perception tokens as chain-of-thought tokens (see \cref{fig:teaser}).
Additionally, we adopt a curriculum learning approach to avoid catastrophic forgetting.


We apply the \ours training algorithm to the LLaVA~\cite{liu2023visualinstructiontuning} model, resulting in our LLaVA-\ours variant. Our LLaVA-\ours model significantly outperforms standard fine-tuning approaches across multiple perception-demanding tasks, demonstrating the generality and effectiveness of our method. LLaVA-\ours achieves state-of-the-art results on both relative depth estimation and object counting tasks. For instance, on BLINK relative depth estimation, LLaVA-\ours delivers a performance boost of $6.4\%$ points compared to the fine-tuning baseline. Similarly, on counting tasks, LLaVA-\ours drives improvements of $10.8\%$ points on BLINK, $11.3\%$ points on CVBench, and $8.3\%$ on SEED-Bench. \cref{fig:image_figure} illustrates examples from these tasks.
Perception tokens open up a whole new modality through which MLMs can begin to reason, tackling tasks beyond just language reasoning.

\section{Related work}
\label{sec:relatedwork}

\paragraph{Multimodal language models (MLMs).} MLMs aim to solve a variety of tasks (e.g., Visual Question Answering (VQA) and captioning) based on vision and language inputs. Most modern architectures accomplish this by relying on either cross-attention~\citep{alayrac2022flamingovisuallanguagemodel, bai2023qwenvlversatilevisionlanguagemodel} or visual instruction tuning~\citep{liu2023visualinstructiontuning, liu2024improvedbaselinesvisualinstruction, li2023blip2bootstrappinglanguageimagepretraining, ye2024mplugowlmodularizationempowerslarge} to interleave multimodal information. Cross-attention architectures operate by independently encoding images and then cross-attending to a language model backbone. On the other hand, visual instruction tuning produces token embeddings from image representations that can be interleaved with language tokens to ground generation. While these techniques work well for high level tasks, MLMs still struggle with mid-level and low-level tasks such as counting, depth reasoning, and segmentation. Most MLMs can be classified as either end-to-end MLMs or tool-using MLMs.

\paragraph{End-to-end MLMs.} End-to-end MLMs use a single, unified architecture~\citep{alayrac2022flamingovisuallanguagemodel, dai2023instructblipgeneralpurposevisionlanguagemodels, li2023mimicitmultimodalincontextinstruction, li2023blip2bootstrappinglanguageimagepretraining, liu2023llava} that can be repeatedly used for different tasks without requiring any architectural changes. While training end-to-end MLMs can be costly due to the need for large amounts of multi-task data, the use of diverse datasets allows end-to-end MLMs to generalize effectively and learn nuanced visual representations.

\paragraph{Tool-using MLMs.} Tool-using MLMs enable LLMs to perform vision tasks by attaching specialized vision modules such as segmentation or captioning networks~\citep{liu2024mmbenchmultimodalmodelallaround, lu2023chameleonplugandplaycompositionalreasoning, wu2023visualchatgpttalkingdrawing, you2023idealgptiterativelydecomposingvision}. Tool-using MLMs use a router to select the optimal vision module to use for a given input. The use of specialized modules allows MLMs to achieve higher performances on many tasks. However, tool-using MLMs are sensitive to errors because they link together several networks, which are each a potential point of failure.

\paragraph{Any-to-any networks.} To address this, recent works~\citep{wang2022ofaunifyingarchitecturestasks, jin2024unifiedlanguagevisionpretrainingllm, pan2024kosmosggeneratingimagescontext, koh2023generatingimagesmultimodallanguage} such as Unified-IO~\citep{lu2022unifiediounifiedmodelvision, lu2023unifiedio2scalingautoregressive} have experimented with shared embedding spaces and visual decoders for vision and language tasks, training MLMs to generate segmentation masks, keypoints, and depth maps. Similarly,~\citep{wu2024nextgptanytoanymultimodalllm} generates special task tokens to route language model representations to diffusion heads for image, video, and audio generation.
To handle complex segmentation scenarios, LISA~\citep{lai2024lisareasoningsegmentationlarge} and GSVA~\citep{xia2024gsvageneralizedsegmentationmultimodal} train language models to generate an additional segmentation token that can be used to produce segmentation masks, grounding the mask in language reasoning.
\\
While these approaches can generate visual outputs, they cannot reason over their own generations to solve related visual perception tasks. In contrast, augmenting MLMs with perception tokens as chain-of-thought tokens enables them to perform visual reasoning directly, providing significant gains in detail-oriented question-answering tasks such as depth estimation and counting.
\begin{figure*}[ht]
    \centering
    \includegraphics[height=6cm]{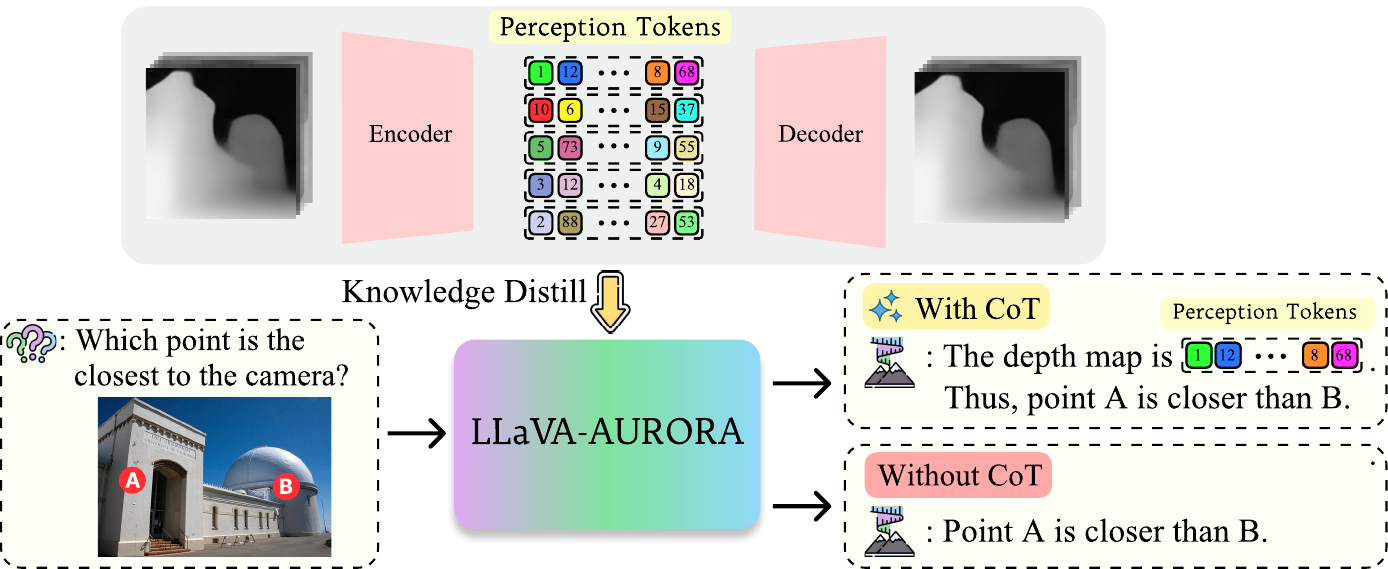}
    \caption{The overall \ours training framework. We first learn visual perception tokens using VQVAE. We then finetune MLMs with a multi-task training approach where we distill intrinsic image representations (\eg, depth map) into MLMs by training them to decode the visual tokens as intermediate reasoning steps towards completing the tasks.}
    \vspace{-4mm}
    \label{fig:aurora}
\end{figure*}

\section{Perception Tokens \& Aurora}
\label{sec:methods}

We introduce \textbf{Perception Tokens} and our \textbf{Aurora} training algorithm (see \cref{fig:aurora}), which augment multimodal language models with perception tokens, enabling the MLM to leverage these tokens effectively during training and incorporate them into its chain-of-thought reasoning process for enhanced visual reasoning.

\subsection{Problem formulation}
In autoregressive large language models, chain-of-thought (CoT) reasoning can be formulated as a multi-step inferential process in which the model iteratively generates intermediate reasoning steps to arrive at a final answer. Given a task input $x$, the model generates a response $y$ conditioned on the input and a sequence of $m$ intermediate reasoning steps $\{s_i\}_{i=1}^{m}$, where $x$, $y$, and each $s_i$ are sequences of tokens from the model's vocabulary $V$. 

Existing multimodal language models often rely on limited vocabulary tokens derived from text or pre-trained image embeddings like CLIP, restricting their capacity to interpret other representations crucial for reasoning. Mid-level and low-level visual features such as depth maps, instance segmentation masks, human poses and bounding boxes which could substantially improve visual reasoning, are currently incompatible with the model and cannot be integrated during training or inference. Our key insight is to introduce auxiliary perception tokens for these intermediate steps with an expanded vocabulary $V'=V \cup V_{\mathrm{aux}}$, bridging this gap by allowing the model to integrate richer visual representations into its reasoning process. Conditioning the final output on these tokens enhances the model's accuracy and interpretability across multimodal tasks.

\subsection{Perception token prediction and reasoning}

Introducing an expanded vocabulary to enhance multimodal reasoning presents two main challenges. The first challenge is enabling the model to generate tokens from the new auxiliary vocabulary $V_{\mathrm{aux}}$, which includes specialized tokens for low- and mid-level visual features, such as depth maps and bounding boxes. The second challenge is ensuring that the model can effectively condition on these auxiliary tokens to improve reasoning, particularly for multi-step inference tasks.

\paragraph{Perception token prediction.} To address the first challenge, we employ a \textit{specialist-to-generalist distillation} approach, using pre-trained specialist models (e.g., depth estimation or instance segmentation) to guide auxiliary token generation through cross-entropy loss. For each input $x$, the specialist model provides a target probability distribution $q_i$ over its tokens. Let $M: V_{\mathrm{spec}} \rightarrow V_{\mathrm{aux}}$ denote a one-to-one mapping from the specialist model's vocabulary $V_{\mathrm{spec}}$ to the auxiliary token vocabulary $V_{\mathrm{aux}}$. We define the distillation loss as:
\begin{equation}
\ell_{dist} = \min_M \big(-\sum_i q_i \log p_{M(i)}\big),
\end{equation}
where $p_{M(i)}$ is the probability assigned by our model to the auxiliary token corresponding to the mapped token $M(i)$. This consistent mapping allows the model to effectively align its predictions with the specialist model’s output distribution, enhancing the relevance and accuracy of its auxiliary token predictions.

In addition to distillation, we incorporate a \textit{reconstruction} loss to enhance the token prediction and the interpretability of our model. Each auxiliary token corresponds to a specific representation, such as a depth map or a bounding box vector, and is trained to directly predict this feature. To achieve this, we introduced a lightweight decoder $g$ that maps the tokens into the feature space, allowing for efficient and interpretable transformations. Formally, for a token $t\in V_{\mathrm{aux}}$, decoder $g$, and its target feature $f$, the reconstruction loss is defined as:
\begin{equation}
\ell_{rec} = \Vert g(t) - f\Vert_2^2,
\end{equation}
where $g(t)$ is the decoded representation for token $t$ in the feature space. This reconstruction process not only aligns each token with a meaningful feature, improving interpretability, but also reinforces the accuracy of auxiliary token predictions through direct feature supervision. In practice, while the reconstruction objective improves performance and interpretability, it is optional, as models can be effectively trained with the distillation objective alone to reduce computational overhead.

\paragraph{Reasoning with perception tokens.} The second challenge is enabling the model to condition on tokens from $V_{\mathrm{aux}}$ effectively when generating each subsequent reasoning step, thereby enhancing its reasoning capabilities. To achieve this, we introduce chain-of-thought reasoning progressively, beginning with simpler, single-step reasoning tasks and advancing to more complex, multi-step inference. The model begins by learning single-step reasoning, predicting an initial reasoning step $s_1$ based on the input $x$. It then progresses to multi-step reasoning, predicting sequences $s_1, \cdots, s_m$ and effectively utilizing auxiliary tokens to support extended chains of inference. We further reinforce this process with \textit{constrained decoding} and an \textit{information bottleneck}: in constrained decoding, we restrict sampling to auxiliary tokens, ensuring they serve as intermediate reasoning steps; in the information bottleneck approach, we truncate the reasoning chain before the auxiliary token when generating subsequent reasoning steps, forcing the model to rely solely on auxiliary tokens to reach the correct answer. Lastly, we provide a multi-task data synthesis approach to train the model using \textbf{curriculum learning} across various synthetic tasks. Further details are provided in Section~\ref{subsec:progressive}.

\subsection{Tokenization}

A unified tokenization space is crucial for multimodal models as it creates a consistent framework through which varied visual tasks can be represented, processed, and interpreted. Inspired by~\cite{ning2023all}, we establish a unified tokenization space which enables the model to enable the model to learn varied visual features within a shared representation seamlessly. In our experiments, we implement two tailored tokenization schemes for commonly used visual representations. Importantly, our framework is designed with flexibility, allowing it to generalize to a broad range of visual representations beyond those presented here.

\paragraph{Pixel-level representation.} This tokenization scheme captures fine-grained spatial information, such as depth maps and segmentation masks, providing the model with detailed pixel-level data essential for accurate visual processing. For these types of tokens, we leverage visual tokenizers like VQVAE and VQGAN, which take in the ground truth masks or depth maps and return discrete target tokens~\cite{van2017neural, esser2021taming, lu2022unifiediounifiedmodelvision, lu2023unifiedio2scalingautoregressive}.

\paragraph{Structured representation.} This scheme encodes structured yet abstracted visual features, such as human poses, bounding boxes, and coordinates, allowing the model to reason with higher-level spatial relationships and object hierarchies. For these tokens, we define the domain of the tokens based on specific properties; for example, the domain for coordinates can range from 0 to the maximum number of pixels in the image’s height or width~\cite{chen2021pix2seq}.

\begin{table*}[t]
\centering

\scalebox{0.8}{
\begin{tabular}{@{}p{3.5cm}p{1cm}p{0.05cm}p{1cm}p{0.05cm}p{1cm}p{0.05cm}p{1.2cm}p{0.1cm}p{1.5cm}p{0.1cm}p{1.5cm}p{0.1cm}p{1.5cm}p{0.1cm}p{1.5cm}p{0.1cm}p{1.5cm}@{}}

\toprule

\multicolumn{1}{c}{} & \multicolumn{5}{c}{\textbf{Training}} & \multicolumn{10}{c}{} \\ 
\cline{2-6}\\
\textbf{Model} & Direct Labeling Data && Depth Generation Data && CoT Data  && BLINK~\cite{fu2024blink} 2 Points &&  HardBLINK 3 Points && HardBLINK 4 Points && HardBLINK 5 Points && Average\\ 

\midrule

LLaVA OneVision & \crossmark && \crossmark && \crossmark && 51.6  && 33.1 &&  22.6 && 18.5 && 31.4 \\
LLaVA 1.5 13B & \crossmark && \crossmark && \crossmark && 54.0  && 35.5 && 37.9 &&  29 && 39.1  \\
Fine-tunned LLaVA & \ourcheckmark && \crossmark && \crossmark && 68.5  && 58.9 &&  52.4 && 41.1 && 55.2  \\
LLaVA-\ours (Ours) & \ourcheckmark && \ourcheckmark && \ourcheckmark &&  64.5  && \textbf{66.9 }&& \textbf{60.5} && \textbf{54.8} && \textbf{61.6} \\

\midrule

\textcolor{gray}{GPT-4o} & \crossmark && \crossmark && \crossmark &&  \textcolor{gray}{53.2} && \textcolor{gray}{58.9} && \textcolor{gray}{50} && \textcolor{gray}{36.3} && \textcolor{gray}{49.6}   \\
\textcolor{gray}{GPT-4 Turbo} & \crossmark && \crossmark && \crossmark && \textcolor{gray}{58.1}  && \textcolor{gray}{54.8} && \textcolor{gray}{41.9} && \textcolor{gray}{32.2} &&  \textcolor{gray}{46.7}  \\
\textcolor{gray}{GPT-4 Turbo + Tool} & \crossmark && \crossmark && \crossmark && \textcolor{gray}{\textbf{70.2}} && \textcolor{gray}{57.2} && \textcolor{gray}{44.3} && \textcolor{gray}{26.6} &&  \textcolor{gray}{49.6}   \\

\bottomrule
\end{tabular}
}

\caption{Performance comparison between our LLaVA-\ours model, the fine-tunning baseline, and the original base model on the relative depth accuracy (\%) task. Results demonstrate that our approach, utilizing depth tokens and intermediate reasoning steps, significantly outperforms both the baseline and the base model, particularly on more challenging configurations with 3, 4, and 5 points sampled from the image's mid-height region.}
\label{tab:relative_depth}
\end{table*}

\subsection{Curriculum learning with progressive CoTs}
\label{subsec:progressive}

The objective of training the model is to develop a data and computationally efficient method for learning to predict novel, fine-grained visual tokens and using them to complete complex visual reasoning tasks. We observed that the standard approach, which relies on a fixed mixture data, encounters a trade-off between the accuracy of novel tokens predictions and the model's reasoning capability, primarily due to catastrophic forgetting and challenges in reasoning with new tokens. Conversely, fine-tuning the model with the original training mixture significantly raises computational costs and may be impractical if the original data, particularly proprietary datasets, is unavailable, making this approach less scalable for incorporating new tokens in the future.

We propose a curriculum learning-inspired training scheme that begins with atomic tasks and gradually advances to more complex ones requiring sophisticated, multi-hop reasoning. Let $d_t$ represent the difficulty of task $t$, with difficulties $d_1 < d_2 < \cdots d_T$ across $T$ tasks, and let $p(t,s)$ denote the probability of sampling data points from task $t$ at training step $s$. We define $p(d_t, s)$ using a temperature-scaled Softmax formulation as follows:
\begin{equation}
p(d_t, s) = \frac{\exp{(-d_t/\tau(s))}}{\sum_{i=1}^m\exp{(-d_i/\tau(s)})},
\end{equation}
where $\tau(s)$ modulates the task difficulty over time, allowing a smooth shift in the probability distribution toward harder tasks. This temperature function is defined as:
\begin{equation}
\tau(s)=\frac{\tau_0}{1+\lambda \cdot s/S}.
\end{equation}
Here, $\tau_0$ is the initial temperature, $\lambda$ is the annealing rate, and $S$ is the total number of training steps.

Our approach for defining $d_t$ values is based on the inherent complexity of each task, which corresponds to the depth of reasoning involved. Specifically, we assign $d_1$ to the most atomic task, involving the prediction of newly introduced tokens. At the other end of the spectrum, $d_m$ represents the final task, requiring the full chain-of-thought (CoT) reasoning steps. Between them, intermediate tasks $\{d_t\}_{t=1}^{T}$, serves to bridge the gap between the atomic tasks and the comprehensive chain-of-thought responses.

In this project, we introduce three types of data subsets for each downstream task, organized in increasing levels of difficulty. The first and most atomic task involves teaching the model to generate tokens from the new auxiliary vocabulary . For instance, in depth-related tasks, we train the model to learn depth maps; in segmentation-related tasks, we teach the model to generate masks; and so on.

The other two data subsets involve Chain-of-Thought (CoT) prompts and direct labeling which help with reasoning with auxiliary tokens. In the CoT subset, we use the new intermediate visual perception tokens to answer downstream-specific questions, encouraging the model to reason step by step. In the direct labeling subset, we pose the same questions but instruct the model to provide direct answers without step-by-step reasoning.

Inspired by \cite{hsieh2023distilling}, we employ a multitasking approach for these two data subsets. For each image, we sequentially present both the CoT and direct labeling questions, allowing the model to tackle each image with both reasoning styles in sequence. We use a sequential sampler rather than a random sampler, shuffling the images beforehand. This strategy enables the model to learn from both types of reasoning tasks effectively, enhancing its ability to perform complex visual reasoning.

\section{Experiments}
\label{sec:experiments}

We base our work on LLaVA 1.5 13B as the foundation for our model, which we refer to as LLaVA-\ours. Our approach augments the MLM with perception tokens to enhance reasoning and improve performance across both 3D and 2D visual tasks. We evaluate our approach on relative depth estimation (3D) using pixel-level depth map tokens for fine-grained depth capture, and on object counting (2D) with mid-level bounding box tokens for precise localization. These tokens not only enhance task-specific results but also highlight our framework’s potential to generalize effectively across a broad spectrum of visual reasoning tasks.

\subsection{3D reasoning task}
We choose relative depth estimation as our 3D task because it enables the model to determine spatial relationships within a scene by identifying which points are closer to or farther from the camera. This foundational skill is crucial for scene understanding and applications requiring spatial awareness, such as robotics and autonomous systems. Specifically, this task involves identifying the point closest to the camera among multiple marked points in an image. To support the model's reasoning, we use discrete depth map tokens that capture spatial depth information, enhancing the model’s understanding of proximity.

\paragraph{Tokenization.} To capture fine-grained spatial details, we tokenize depth maps into sequences of discrete tokens. Inspired by the approach in AiT~\cite{ning2023all}, we use a Vector Quantized Variational Autoencoder (VQVAE)~\cite{van2017neural} with a codebook size of 128. In this setup, each depth map is encoded as a grid of embeddings, with each embedding matched to the nearest entry in the codebook, yielding a compact depth representation. The VQVAE decoder reconstructs the depth map from this sequence of latent codes, and the entire model is optimized with a reconstruction loss to ensure precise encoding. During inference, each 320x320 depth map is compressed into a 10x10 grid of code indices, resulting in a 100-token sequence where each token represents one of the 128 discrete depth tokens, labeled $\texttt{DEPTH\_0}$ to $\texttt{DEPTH\_127}$. To organize the sequence, we encapsulate it with special tokens $\texttt{DEPTH\_START}$ and $\texttt{DEPTH\_END}$, adding a total of 130 depth-related tokens to the model’s vocabulary.

\begin{table*}[ht]
\centering

\scalebox{0.8}{
\begin{tabular}{@{}p{3.5cm}p{1.5cm}p{0.05cm}p{1cm}p{0.05cm}p{1.5cm}p{0.05cm}p{1.5cm}p{0.1cm}p{2cm}p{0.1cm}p{1.5cm}p{0.1cm}@{}}
\toprule

\multicolumn{1}{c}{} & \multicolumn{5}{c}{\textbf{Training}} & \multicolumn{6}{c}{} \\ 
\cline{2-6} 
\textbf{Model} & Direct Labeling Data && Bounding Box Data && CoT Data  && CV-Bench Counting && SEED-Bench Counting  && BLINK Counting \\ 

\midrule

LLaVA One Vision & \crossmark && \crossmark && \crossmark && 34.4  && 31.7 && 35.8   \\
LLaVA 1.5 13B & \crossmark && \crossmark && \crossmark &&  40.9 && 52.2 && 35.0  \\
Fine-tunned LlaVA & \ourcheckmark && \crossmark && \crossmark && 44.7 && 46.3  && 0.2     \\
LLaVA-\ours (Ours) & \ourcheckmark && \ourcheckmark && \ourcheckmark && 56.0 && 54.6  &&  45.8   \\

\midrule

\textcolor{gray}{GPT-4o} & \crossmark && \crossmark && \crossmark && \textcolor{gray}{70.18} &&  \textcolor{gray}{64.6}  && \textcolor{gray}{47.5} \\
\textcolor{gray}{GPT-4 Turbo} & \crossmark && \crossmark && \crossmark &&  \textcolor{gray}{61.3}&& \textcolor{gray}{64.8} && \textcolor{gray}{57.5} \\
\textcolor{gray}{GPT-4 Turbo + Tool} & \crossmark && \crossmark && \crossmark &&  \textcolor{gray}{48.6} && \textcolor{gray}{29.9} && \textcolor{gray}{26.7} \\

\bottomrule
\end{tabular}
}
\caption{Comparison of object counting accuracy (\%) across three benchmarks (CV-Bench, SEED-Bench, and BLINK). Our LlaVA-\ours model, using auxiliary perception tokens to encode bounding box information for intermediate reasoning, demonstrates superior performance compared to the fine-tunning baseline models and the original base model.}
\label{tab:counting}
\end{table*}
\vspace{-0.05em}
\paragraph{Training data.} We train the VQVAE model on pseudo-depth maps generated from the ADE20k dataset~\cite{zhou2017scene, zhou2019semantic} using the Depth Anything model~\cite{yang2024depth, yang2024depthv2}. This dataset provides a diverse range of scenes, enhancing the model's ability to generalize.

For fine-tuning, we prepare three types of data tailored for relative depth estimation (as detailed in \cref{subsec:progressive}). First, we generate depth maps for 20k ADE20k images, tokenize them with the pre-trained VQVAE, and format each sample in a Q\&A structure, prompting the model with a depth estimation question and an answer sequence of depth tokens. Additionally, we construct a dataset of 500 ADE20k images for chain-of-thought (CoT) reasoning, with 2–5 markers in each image. Here, the prompt guides the model to generate the coordinates of the markers and then the depth map as intermediate steps, then identify the marker closest to the camera. This CoT training improves sequential reasoning and relative depth estimation accuracy. Finally, we use the same 500 images for direct labeling, prompting the model to directly identify and label the marker closest to the camera (More details in Supplementary).

This fine-tuning setup enables our model to effectively handle  step-by-step reasoning for the relative depth estimation task.







\subsection{2D reasoning task}
We select counting as a critical 2D visual task. For object counting, we incorporate bounding box predictions as an intermediate reasoning step to improve accuracy in answering counting queries. Given an image and a question about the number of specific objects, the model first identifies and predicts bounding boxes for each instance of the target object. These bounding box tokens serve as structured, intermediate representations, enabling the model to understand spatial arrangements and accurately count object instances.
\vspace{-1.5em}
\paragraph{Tokenization.} To represent bounding boxes as discrete tokens, we resize all input images to a fixed resolution of 336x336 pixels. This preprocessing step allows us to add 336 unique tokens to the model’s vocabulary, each representing a specific pixel position within the resized image. These tokens, labeled \texttt{PIXEL\_0} to \texttt{PIXEL\_335}, enable the model to uniquely reference each pixel location. Bounding boxes are encoded as tuples of four tokens, formatted as (\texttt{PIXEL\_i}, \texttt{PIXEL\_j}, \texttt{PIXEL\_k}, \texttt{PIXEL\_m}), where \texttt{PIXEL\_i} and \texttt{PIXEL\_j} denote the coordinates of the top-left corner and \texttt{PIXEL\_k} and \texttt{PIXEL\_m} represent the bottom-right corner (i.e., $(x_1, y_1, x_2, y_2)$). This discrete representation allows the model to interpret and use bounding box locations effectively, providing the spatial structure needed for accurate object counting.

\paragraph{Training data.} To fine-tune the model for object counting, we draw on three types of data tailored to this task (as detailed in \cref{subsec:progressive}). First, we use task-specific data from the LVIS dataset~\cite{gupta2019lvis}, selecting 5k images with objects whose counts range from 0 to 15. For each selected image, we specify an object type (e.g., "beds") and structure the fine-tuning samples to prompt the model for bounding box predictions of the specified objects within the image.

To enhance the model’s reasoning ability, we include a small subset of 250 LVIS images for chain-of-thought (CoT) training. Here, each question prompt encourages step-by-step reasoning by instructing the model to first generate bounding boxes for the target object, followed by providing the final count. Additionally, we create a direct labeling subset using the same 250 images. In this subset, we prompt the model to directly identify and label the total count of the specified objects without the intermediate step of bounding box generation.

\subsection{Benchmarks}

\paragraph{Relative depth.} A recent benchmark, BLINK~\cite{fu2024blink}, introduces tasks designed to be intuitive for humans yet challenging for multimodal models, with relative depth estimation as one of its tasks. BLINK provides 124 images, each containing two marked points labeled as $A$ and $B$, and asks which point is closer to the camera. To reduce biases that language models have toward answering multiple-choice questions~\cite{zheng2023large,pezeshkpour2023large}, we modify the original BLINK questions by removing the answer choices. 
 To further evaluate the model’s reasoning and 3D understanding in relative depth, we curated a series of more challenging benchmark sets, collectively called HardBLINK. We progressively increase task difficulty by altering the prompts and image configurations as follows:
 \begin{enumerate}
     \item Prompt Modification: In the prompts, we exclude the number of markers and their labels, requiring the model to infer these details solely from the image.
     \item Increased Point Complexity: We generate four variations of BLINK by adding more markers to each image, using Depth Anything to produce pseudo-depth maps for precise placement. These curated sets— HardBLINK 3points, HardBLINK 4points, and HardBLINK 5points—contain the same images as BLINK but with 3, 4, and 5 randomly placed markers, respectively, each with reasonable depth and distance differences. This setup tests the model’s depth reasoning across more complex spatial configurations.
     \item Mitigating Height-Based Bias: To prevent the model from assuming that higher points are farther from the camera, we place markers at mid-height within the image. This approach encourages reliance on depth information rather than positional cues~\cite{chen2016single}.
 \end{enumerate}
\paragraph{Counting.} For counting, we evaluate the model on CV-Bench~\cite{tong2024cambrian1}, SEED-Bench~\cite{li2023seed}, and BLINK’s counting~\cite{fu2024blink} subtask. To better capture the model’s capabilities, we also remove multiple-choice options, requiring it to generate an exact count.

\begin{table*}[ht]
\centering

\scalebox{0.9}{
\begin{tabular}{@{}lp{1.2cm}p{0.1cm}p{1.5cm}p{0.1cm}p{1.5cm}p{0.1cm}p{1.5cm}p{0.1cm}p{1.5cm}p{0.1cm}p{1.5cm}@{}}

\toprule

\multirow{2}{*}{\textbf{Model}} & BLINK~\cite{fu2024blink} 2 points  && HardBLINK 3 Points && HardBLINK 4 Points && HardBLINK 5 Points && Average \\ 

\midrule

VQGAN~\cite{esser2021taming} (16384 codes) & 82.2  && 66.1 && 53  && 37 && 59.6  \\
Unified-IO & 70.2  && 75.8 && \textbf{75.8} &&  \textbf{75.8} && 74.4\\
Unified-IO2 & 54  && 37.9 && 21 && 27.4  && 35.1\\
LLaVA-\ours (Ours) & \textbf{91.9}  &&  \textbf{78.2} &&  71.7&& \textbf{75.8}  &&  \textbf{79.2}  \\
\hline
\textcolor{gray}{Our VQVAE (128 codes)} & \textcolor{gray}{96.7}   && \textcolor{gray}{94.3} && \textcolor{gray}{95.2} && \textcolor{gray}{96.7} && \textcolor{gray}{95.7}\\

\bottomrule
\end{tabular}
}

\caption{While not the main aim of our work, we report the \textbf{depth generation performance} across benchmarks with 2, 3, 4, and 5 marked points using BLINK~\cite{fu2024blink}'s relative depth subtask images. We report relative depth estimation accuracy (\%), calculated by programmatically extracting depth values at specific coordinates from model-generated depth maps. Our model consistently outperforms other multimodal models, including Unified-IO~\cite{lu2022unifiediounifiedmodelvision} and Unified-IO 2~\cite{lu2023unifiedio2scalingautoregressive}}
\label{tab:depth}
\end{table*}
\begin{figure}[h!]
    \centering
    \includegraphics[width=\columnwidth]{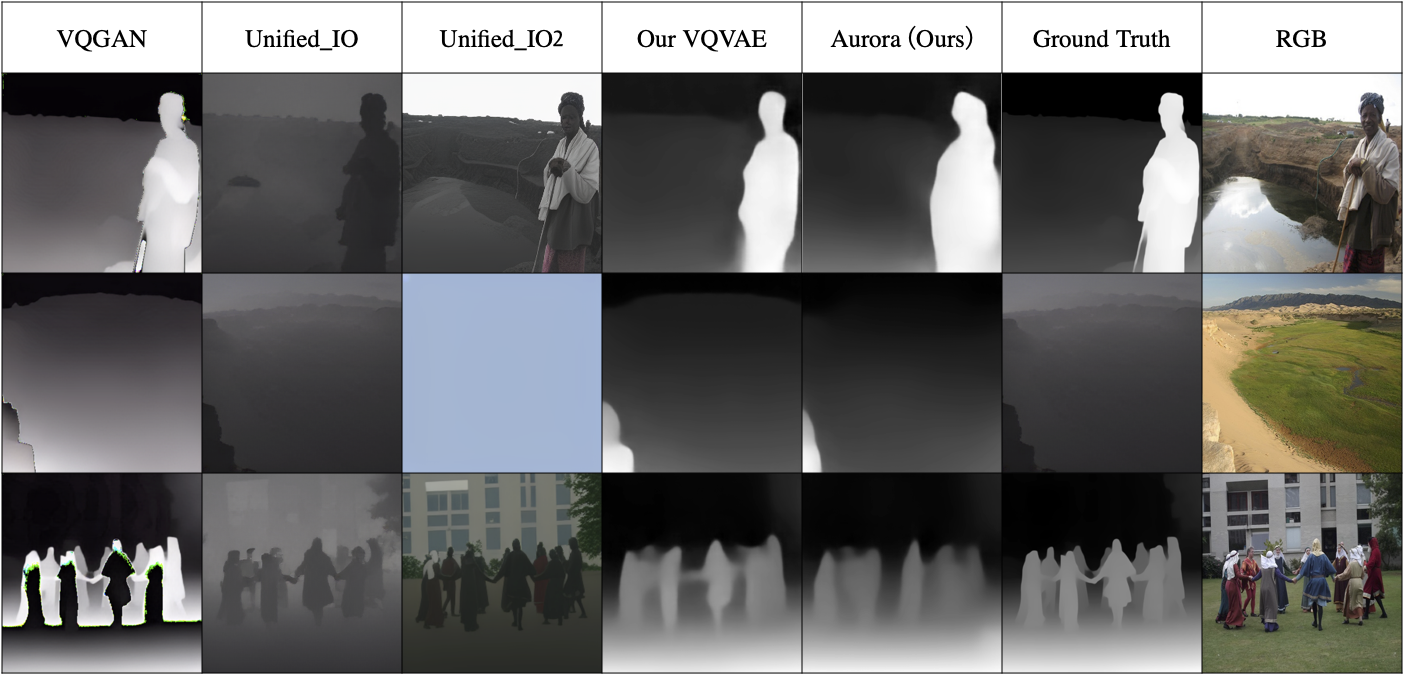}
    \caption{Depth maps generated by Aurora are imperfect but resemble the ground-truths from Depth Anything~\cite{yang2024depth}. }
    \label{fig:depth_vis}
\end{figure}


\subsection{Baselines}
We evaluate a diverse range of models, including closed-source models like GPT-4o~\cite{achiam2023gpt} and GPT-4 Turbo~\cite{achiam2023gpt}, as well as state-of-the-art open-source models such as LLaVA OneVision~\cite{li2024onevision}. Another key baseline for our work involves fine-tuning the base model \ours is applied on, solely on the direct labeling portion of the training data for each task, omitting the newly introduced tokens. This approach allows us to isolate the impact of our token-based enhancements. Additionally, we evaluate the base model for \ours, LLaVA 1.5 13B, to assess its performance without task-specific adaptations. For comparison, we use a tool-augmented baseline with an LLM. 

In relative depth estimation, we employ GPT-4 Turbo, providing it with the ground truth depth maps generated by Depth Anything for each image, allowing it to use this information in its responses. Details on the exact format are provided in the supplementary materials~\cite{hu2024sketch}. For counting, we also use GPT-4 Turbo in a tool-augmented setup. In this process, GPT-4 Turbo first identifies the object specified in the question, then uses Grounding DINO~\cite{liu2023grounding} to locate the bounding boxes for each instance of the object and finally counts them.

\subsection{How new tokens improve 3D reasoning?}
Our experiments demonstrate that incorporating new visual tokens significantly enhances the model’s 3D reasoning abilities, specifically in the relative depth estimation task. Results in \cref{tab:relative_depth} show that our model outperforms both the primary baseline, which is fine-tuned solely on direct labeling data, and the original base model, indicating that the added tokens contribute meaningfully to task accuracy. 

As task complexity increases—for instance, when additional markers are introduced in the HardBLINK benchmarks—our model’s performance advantage becomes even more pronounced. Not only does it maintain high accuracy in distinguishing depth relationships, but it also surpasses both advanced closed-source models, such as GPT-4 Turbo, and GPT-4 Turbo + Tool that use ground truth depth maps. This suggests that the new tokens enable our model to develop a more nuanced understanding of depth cues, even in cases where complex spatial reasoning is required.

Overall, these results highlight the effectiveness of our token-based approach in enhancing 3D reasoning, allowing the model to handle increasingly difficult tasks with robust performance and accuracy.

\subsection{How new tokens improve 2D reasoning?}
In the 2D task of object counting, incorporating new visual tokens provides a significant advantage over the primary baselines. As shown in \cref{tab:counting}, our model outperforms both the baseline fine-tuned solely on direct labeling data and the original base model, as well as the state-of-the-art open-source model LLaVA-OneVision and GPT-4 Turbo + Tool, underscoring the value of these tokens in enhancing counting accuracy.

Although our model does not yet surpass advanced closed-source models, the results demonstrate that the new tokens yield a meaningful improvement in 2D reasoning, enabling more reliable and accurate object counting compared to standard fine-tuning approaches and open-source alternatives.

\subsection{Perception token decoding}
Our approach enables the decoding of learned perception tokens into specialist features, such as depth maps and object bounding boxes, to assess their fidelity and utility.
We, in particular, evaluate the accuracy and correctness of the depth maps generated by LLaVA-\ours, specifically focusing on its ability to represent spatial relationships in visual scenes. For this evaluation, we use the relative depth images from the BLINK benchmark~\cite{fu2024blink}, ensuring consistency with our relative depth assessments. LLaVA-\ours generates depth tokens from BLINK images, which are then reconstructed into full depth maps via the decoder of our pre-trained VQVAE model. Notably, the depth maps output by our pre-trained VQVAE provide an upper bound on the quality of the depth maps generated by LLaVA-\ours.
We use programmatic relative depth accuracy as our metric, which measures how well the reconstructed depth maps capture the relative depth of marked points. This evaluation spans several configurations in the benchmark, including images with 2, 3, 4, and 5 labeled points. By comparing the model’s predicted depth with the ground-truth marker coordinates, we calculate relative depth accuracy, allowing us to assess the precision of depth map generation in reflecting spatial depth relationships.
As shown in \cref{tab:depth}, LLaVA-\ours outperforms Unified-IO~\cite{lu2022unifiediounifiedmodelvision,lu2023unifiedio2scalingautoregressive} in this task. Furthermore, qualitative analysis in \cref{fig:depth_vis} reveals that LLaVA-\ours’s depth maps capture spatial details effectively highlighting its capacity to interpret and represent fine-grained depth information.

\section{Conclusion}
\label{sec:conclusion}

Our algorithm enables the lightweight and scalable integration of perception tokens, such as depth maps and bounding box coordinates, into MLMs, allowing them to perform intermediate reasoning steps akin to chain-of-thought processes. Our method achieves state-of-the-art results on challenging tasks like 2D object counting and 3D relative depth estimation. It also enhances model generalization and interpretability without relying on external tools or task-specific finetuning. The framework is inherently adaptable, incorporating new perception tokens as they emerge, making it a future-proof solution for advancing multimodal reasoning.

\noindent\textbf{Acknowledgements.} This work is partially supported by Amazon Science. We also thank Yushi Hu and Lindsey Li for their insightful comments and suggestions.
\newpage
{
    \small
    \bibliographystyle{ieeenat_fullname}
    \bibliography{main}
}
\clearpage
\setcounter{page}{1}
\maketitlesupplementary

\section{Ablation study}
\label{sec:ablations}
In this section, we analyze the impact of various design choices and data configurations on the performance of our proposed method. We focus on three aspects: (1) the impact of including or excluding specific steps in the chain-of-thought (CoT) reasoning process for the 3D task of relative depth estimation, (2) the use of standard text tokens versus new perception pixel tokens for the 2D task of object counting, and (3) the effect of incorporating a perception token reconstruction loss during fine-tunning.

\subsection{Chain-of-thought steps}
For our 3D task of relative depth estimation, the chain-of-thought (CoT) questions in the fine-tunning data include two steps: (1) identifying the coordinates or locations of the points marked in the image, and (2) generating the depth map and determining which point is closer to the camera based on pixel values in the depth map. This study evaluates the impact of including or excluding these steps in the question prompts during fine-tunning.

We experiment with three variations of fine-tunning data configurations:
\begin{enumerate}
    \item Direct Labeling Baseline: The model is fine-tuned solely on direct labeling data, where the question prompts directly ask which point is closer to the camera and provide the label as the answer. These prompts do not include either step (1) or step (2), see baselines section. 
    \item Step (2) Only: This model is fine-tuned with prompts that exclude step (1) (point location identification) but include step (2), asking the model to answer based on the depth map alone.
    \item \ours: Our proposed \ours technique uses prompts that include both steps (1) and (2), explicitly guiding the model through point location identification before generating the depth map.
\end{enumerate}
All models are evaluated on the harder BLINK datasets we introduced. As shown in \cref{tab:cot}, the results demonstrate that having both steps in the prompts provides the most significant performance improvement. This suggests that guiding the model through a multi-step reasoning process in the prompts enables it to better capture spatial relationships and achieve more accurate depth estimations.
\begin{table*}[t]
\centering

\begin{tabular}{p{3.5cm}p{2cm}p{0.05cm}p{2cm}p{0.05cm}p{1.5cm}p{0.1cm}p{1.5cm}p{0.1cm}p{1.5cm}p{0.1cm}p{1.5cm}}

\toprule

\multicolumn{1}{c}{} & \multicolumn{3}{c}{\textbf{CoT steps}} & \multicolumn{6}{c}{} \\ 
\cline{2-4}\\
\textbf{Model} &   Coordinates && Depth   &&  HardBLINK 3 Points && HardBLINK 4 Points && HardBLINK 5 Points \\ 

\midrule

Direct Labeling Baseline & \crossmark &&  \crossmark  && 58.9 &&  52.4 && 41.1   \\
Step (2) only & \crossmark && \ourcheckmark  && 56.4 && 56.4 &&  50  \\
LLaVA-\ours (Ours) & \ourcheckmark && \ourcheckmark && \textbf{66.9 }&& \textbf{60.5} && \textbf{54.8}  \\

\bottomrule
\end{tabular}

\caption{Performance comparison of models trained with different Chain of Thought question prompt variations for relative depth estimation on the harder BLINK datasets. Models with both steps in the prompts (\model) achieve the best performance.}
\label{tab:cot}
\end{table*}

\subsection{Text tokens vs. Perception tokens}
In this ablation study, we evaluate the impact of using perception tokens compared to standard text tokens for the object counting subtask. Perception tokens are represented in the format \texttt{PIXEL\_X}, where \texttt{X} is a number between 0 and 336, indicating pixel locations for object bounding boxes. For comparison, we replace these perception tokens with regular text tokens in the fine-tunning data, such that \texttt{PIXEL\_100} is replaced with \texttt{100}, and so on.

As shown in \cref{tab:token}, models utilizing perception tokens achieve higher performance across all three counting benchmarks: BLINK~\cite{fu2024blink}, SEED-Bench~\cite{li2023seed}, and CV-Bench~\cite{tong2024cambrian1}. This demonstrates the effectiveness of perception tokens in explicitly encoding spatial information for improved counting accuracy.

\begin{table*}[h!]
\centering

\scalebox{0.8}{
\begin{tabular}{@{}p{3.2cm}p{0.1cm}p{2cm}p{0.1cm}p{1.5cm}p{0.1cm}p{2cm}p{0.1cm}p{1.5cm}}
\toprule

\textbf{Model} && Token Type&& CV-Bench Counting && SEED-Bench Counting  && BLINK Counting \\ 

\midrule

LLaVA-\ours && Standard  && 52.2 && 50.6  &&  38.3  \\
LLaVA-\ours && Perception  && \textbf{56.0} && \textbf{54.6 } &&  \textbf{45.8}   \\

\bottomrule
\end{tabular}
}

\caption{Comparison of model performance using perception tokens and standard tokens for the object counting task across three benchmarks: BLINK, SEED-Bench, and CV-Bench. Perception tokens consistently improve accuracy.}
\label{tab:token}
\end{table*}

\subsection{Perception token reconstruction loss}

The aim of this ablation study is to assess whether adding the perception token reconstruction loss, despite its increased computational cost, significantly improves model performance. Incorporating this loss requires adding the decoder for the specific task, which increases computation time and resource requirements. Not using it makes the system lighter and faster by just using the token classification loss. Therefore, we evaluate whether the performance gains justify the additional overhead.

To this end, we fine-tune two models based on LLaVA 1.5 13B~\cite{liu2023llava} using a dataset of 20,000 samples only for depth map generation. Each sample includes a prompt such as \textit{"What is the depth map for the image?"} and a response containing sequences of depth tokens. Both models are fine-tuned for 10 epochs: one with the reconstruction loss and one without it (both with cross entropy loss).

The reconstruction loss is computed as the mean squared error (MSE) between the ground truth depth map, which is the output of the VQVAE decoder when provided with the ground truth depth tokens, and the predicted depth map, which is generated by decoding the depth tokens produced by the LLM. A soft merging technique is used in reconstruction, where a "soft token" is created by averaging the embeddings of all potential tokens, weighted by their prediction probabilities from the LLM. 

The models are evaluated on two datasets: (1) 124 images from the relative depth subtask of BLINK~\cite{fu2024blink}, and (2) 1000 random images from the Visual Genome dataset~\cite{krishna2017visual}, for which depth maps were generated using Depth Anything~\cite{yang2024depth}. The evaluation metric is the mean squared error (MSE) between the ground truth decoded depth maps and the depth maps reconstructed from the model's output tokens.

\begin{figure}[h!]
    \centering
    \includegraphics[width=\columnwidth]{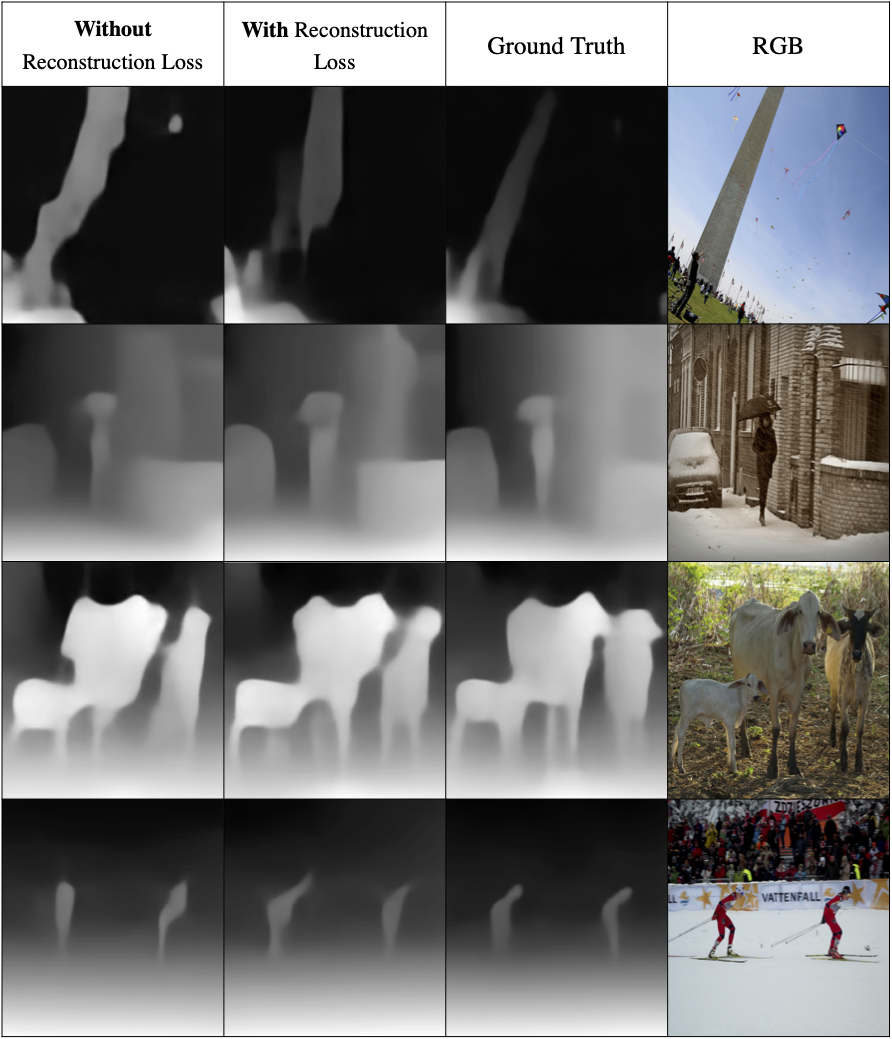}
    \caption{ Qualitative comparison of predicted depth maps with and without reconstruction loss.}
    \label{fig:recons}
\end{figure}

As shown in \cref{tab:recons}, the results indicate that incorporating the reconstruction loss does not significantly improve model performance. \cref{fig:recons} further illustrates qualitative results, highlighting the visual differences in the predicted depth maps with and without the reconstruction loss. While the reconstruction loss enforces consistency between the generated and ground truth depth tokens, its overall contribution is marginal in this setup. This study suggests that omitting the reconstruction loss may be a more efficient choice, especially when computational cost is a concern. Future work could explore its impact in larger datasets or more complex tasks to better understand its potential benefits.

\begin{table}[h!]
\centering

\begin{tabular}{p{1.7cm}p{0.1cm}p{0.5cm}p{0.1cm}p{1cm}p{0.1cm}p{1cm}p{0.1cm}}

\hline
\multicolumn{4}{c}{} & \multicolumn{4}{c}{\textbf{Mean Squared Error$\downarrow$}}  \\ 
\cline{5-8}
Model&& Recons Loss&& BLINK && Visual Genome\\
\hline
LLaVA 1.5 && \ourcheckmark && 0.092 && \textbf{0.074}\\
\hline
LLaVA 1.5&& \crossmark && \textbf{0.087} && 0.076\\
\hline
\end{tabular}

\caption{MSE evaluation of models with and without reconstruction loss on subsets BLINK and Visual Genome datasets.}
\label{tab:recons}
\end{table}

\section{Cross-task generalization}
To assess the generalizability of \ours trained on depth generation and Chain of Thought (CoT) data for the relative depth task, we evaluate it on a different depth-related task. Specifically, we use the Depth subtask from CV-Bench~\cite{tong2024cambrian1}, which involves identifying which of two objects, highlighted with red and blue bounding boxes, is closer to the camera. Similar to the BLINK evaluations for relative depth, we remove options from question prompts in these evaluations too.

As shown in \cref{tab:cross}, our model outperforms both the base LLaVA 1.5 13B model and the fine-tuning baseline, demonstrating its generalization capabilities across depth-related tasks.
\begin{table}[h]
\centering

\begin{tabular}{@{}p{3cm}c}

\toprule

\textbf{Model} &  CV-Bench Depth \\ 

\midrule

LLaVA 1.5 13B &  62.2 \\
Fine-tunned LLaVA & 60.0 \\
\model (Ours) &  \textbf{64.8} \\

\bottomrule
\end{tabular}

\caption{Performance comparison on the CV-Bench Depth subtask, highlighting our model's generalization ability.}
\label{tab:cross}
\end{table}
\section{Implementation details}

\paragraph{Computation resources.} We train Aurora models on single-node machines equipped with 8 A40 GPUs. Each training run completes in less than 10 hours.

\paragraph{Model architecture and token expansion.} Our approach builds on the LLaVA 1.5 13B model~\cite{liu2024improved}, a pre-trained multimodal language model. To support depth-related tasks, we expand the tokenizer by introducing 130 tokens for depth maps and 336 tokens for bounding box coordinates, increasing the vocabulary size beyond the original 32,000 tokens. These additions require modifications to the token embedding layer (embed\_tokens) and the language model head (lm\_head) to accommodate the new tokens.

\paragraph{Fine-tuning approach.} We apply LoRA~\cite{hu2021lora} to the language model for efficient fine-tuning. The vision backbone is kept frozen while the embed\_tokens and lm\_head layers are fully trained. This strategy enables the model to integrate depth and bounding box information without overwriting its pre-trained knowledge.

We fine-tune the model for 10 epochs, using the same LoRA parameters and learning rates as LLaVA. Fine-tunning follows a cross-entropy loss for next-token prediction, treating the new tokens identically to the original vocabulary.

\paragraph{Inference and decoding.}
During inference, we use a temperature of 0 for deterministic generation and employ constrained decoding techniques~\cite{geng2023grammar, Tran-Thien_2024, beurer2024guiding, constrainedwebsite}. For depth map generation, the model outputs exactly 100 depth tokens between $ \texttt{DEPTH\_START}$ and $ \texttt{DEPTH\_END}$, ensuring consistent and structured results.

\paragraph{Curriculum learning for reasoning.}
To enhance the model’s reasoning capabilities, we employ progressive chain-of-thought (CoT) for curriculum learning. Fine-tuning starts with atomic tasks, such as depth map estimation and bounding box predictions, and gradually incorporates multi-tasking data, including CoT reasoning and direct labeling tasks. For instance, in the depth-related task, we use 20,000 samples for depth generation and 1,000 multi-tasking samples (comprising 500 unique images with sequential CoT and direct labeling questions).

In the first epoch, the model is fine-tuned exclusively on 20,000 depth generation samples. Starting from the second epoch, we introduce multi-tasking data by mixing 18,000 random depth generation samples with 2,000 multi-tasking samples (the 1,000 multi-tasking samples repeated twice). This ratio is progressively adjusted in subsequent epochs, culminating in the 10th epoch, where the model is exposed to 2,000 depth generation samples and 18,000 multi-tasking samples. This staged approach ensures a smooth transition from basic tasks to complex reasoning, effectively reinforcing the model's ability to handle multi-step reasoning challenges.

\paragraph{Fine-tuning data.} As discussed in the Methods section, each task is supported by three sub-datasets. For the depth task, these include (1) depth generation data, (2) chain-of-thought (CoT) reasoning data, and (3) direct labeling data. Similarly, for the counting task, the sub-datasets consist of (1) bounding box prediction data, (2) CoT reasoning data, and (3) direct labeling data.

\begin{figure*}[!htbp]
    \centering
    \includegraphics[width=18cm]{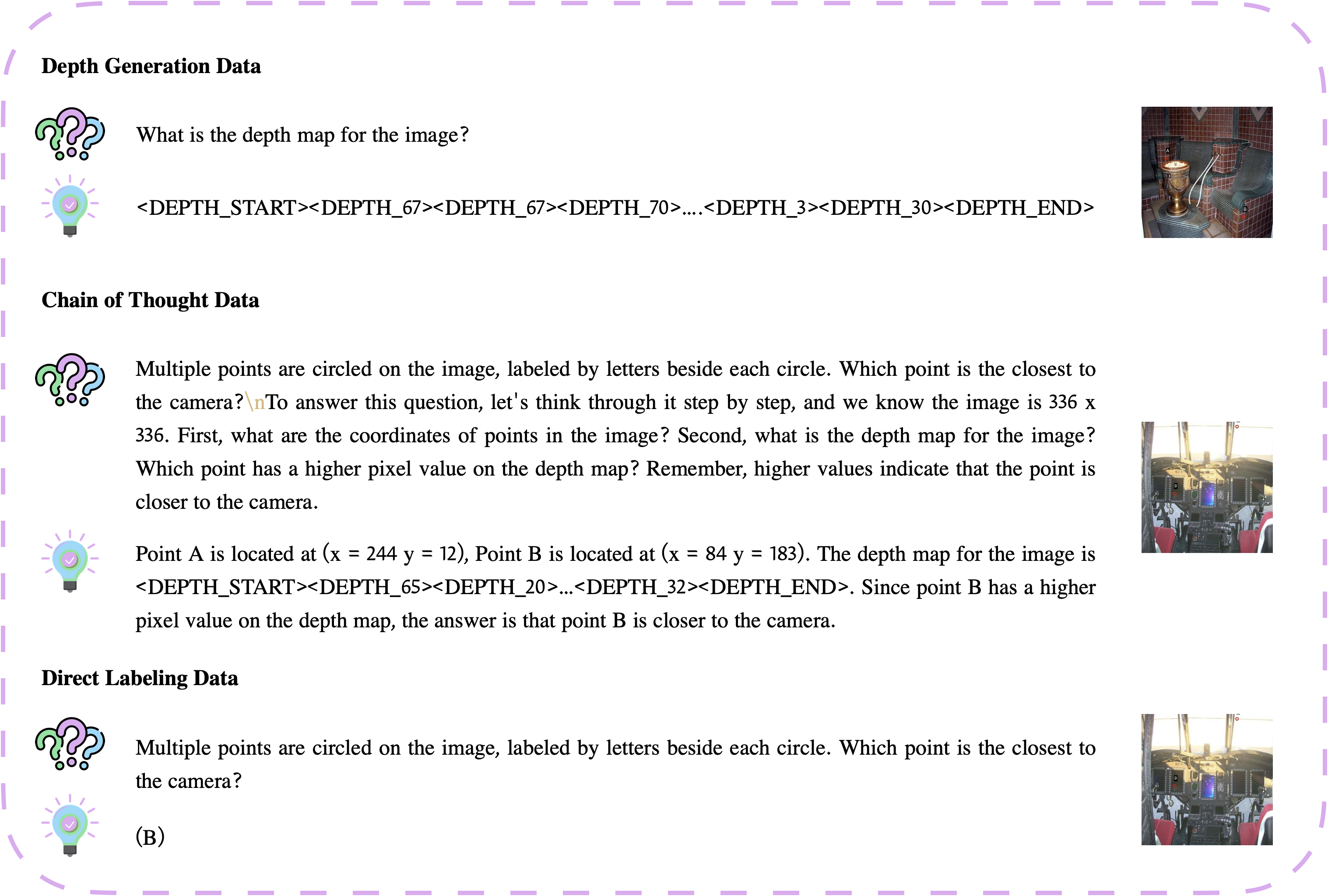}
    \caption{Examples of sub-datasets for the depth task: (1) depth generation, (2) Chain-of-Thought reasoning, and (3) direct labeling.}
    \label{fig:depthdata}
\end{figure*}

\begin{figure*}[!htbp]
    \centering
    \includegraphics[width=18cm]{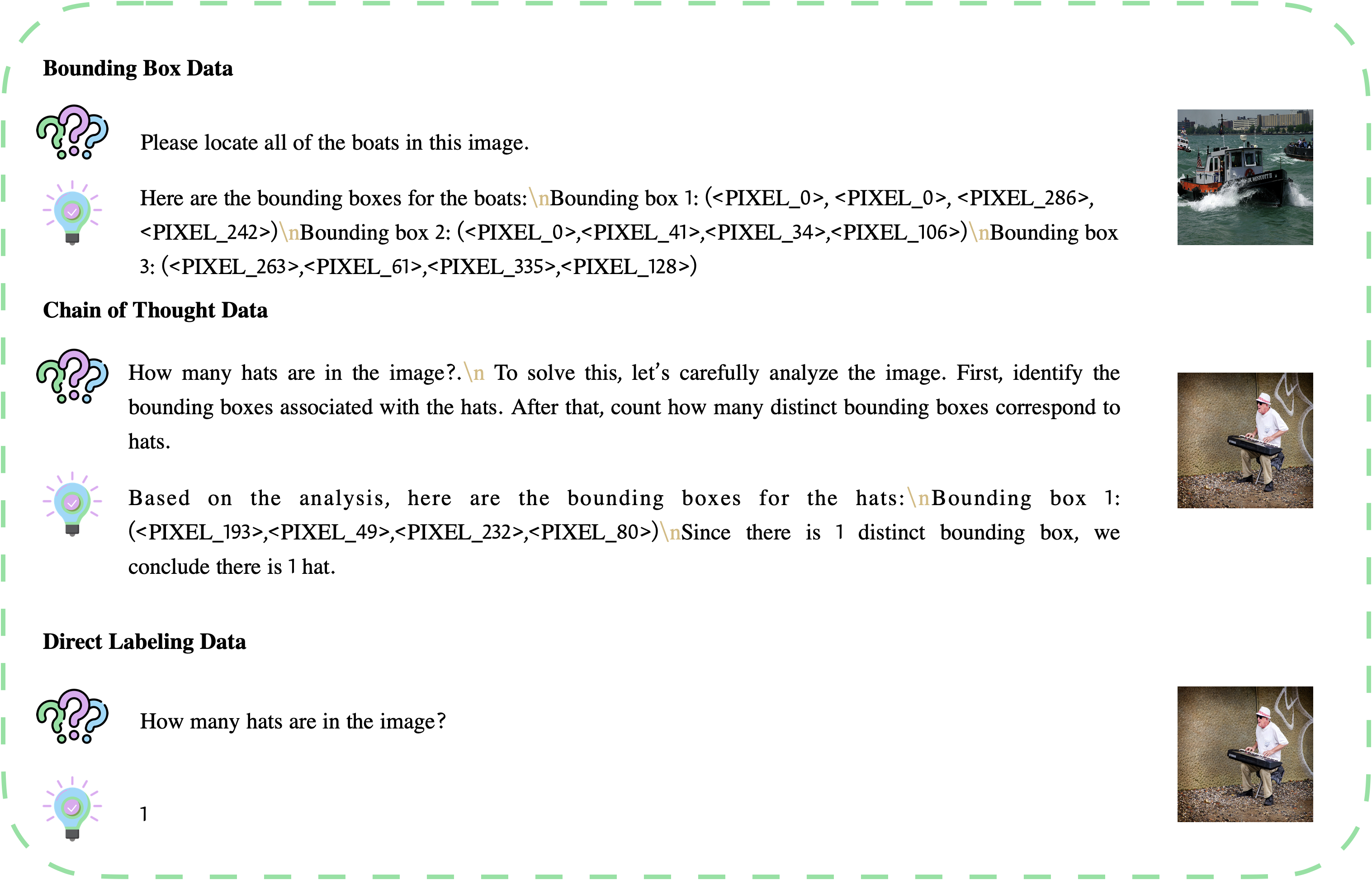}
    \caption{Examples of sub-datasets for the counting task: (1) bounding box prediction, (2) Chain-of-Thought reasoning, and (3) direct labeling.}
    \label{fig:countdata}
\end{figure*}

\cref{fig:depthdata} and \cref{fig:countdata} present representative samples from each sub-dataset for the depth and counting tasks, respectively. 



\end{document}